\documentclass[letterpaper]{article} 
\usepackage{aaai2026}  
\usepackage{times}  
\usepackage{helvet}  
\usepackage{courier}  
\usepackage[hyphens]{url}  
\usepackage{graphicx} 
\usepackage{natbib}  
\usepackage{caption} 
\usepackage{algorithm}
\usepackage{algorithmic}

\usepackage{newfloat}
\usepackage{listings}
\DeclareCaptionStyle{ruled}{labelfont=normalfont,labelsep=colon,strut=off} 
\floatstyle{ruled}
\newfloat{listing}{tb}{lst}{}
\floatname{listing}{Listing}
\usepackage{amsmath}
\usepackage{amssymb}
\usepackage{graphicx}
\usepackage{xcolor}
\usepackage{enumitem}
\usepackage{multirow}
\usepackage{subcaption}
\usepackage{tabularx}
\usepackage{makecell}
\usepackage{pifont}
\usepackage{fontawesome}
\usepackage{threeparttable}
\usepackage{graphicx}
\usepackage{booktabs}

\newcommand{\nop}[1]{}

\title{\sysname{}: Multi-Agent Reinforcement Learning for Incremental DAG Discovery}
\newcommand{\sysname}{MARLIN}
\newcommand{\sysnamevariant}{MARLIN-M}
\newcommand{\sysnameablation}{MARLIN-S}
\author {
    Dong Li\textsuperscript{\rm 1},
    Zhengzhang Chen\textsuperscript{\rm 2}\thanks{Corresponding authors}, 
    Xujiang Zhao\textsuperscript{\rm 2},
    Linlin Yu\textsuperscript{\rm 3},
    Zhong Chen\textsuperscript{\rm 4},\\
    Yi He\textsuperscript{\rm 5},
    Haifeng Chen\textsuperscript{\rm 2},
    Chen Zhao\textsuperscript{\rm 1*}
}
\affiliations {
    \textsuperscript{\rm 1}Department of Computer Science, Baylor University\\
    \textsuperscript{\rm 2}NEC Labs America\\
    \textsuperscript{\rm 3}School of Computer and Cyber Sciences, Augusta University\\
    \textsuperscript{\rm 4}School of Computing, Southern Illinois University\\
    \textsuperscript{\rm 5}Department of Data Science, The College of William and Mary\\
    \{dong\_li1, chen\_zhao\}@baylor.edu,
  \{zchen, haifeng\}@nec-labs.com
}

\usepackage{bibentry}

\begin{document}

\maketitle

\begin{abstract}
Uncovering causal structures from observational data is crucial for understanding complex systems and making informed decisions. While reinforcement learning (RL) has shown promise in identifying these structures in the form of a directed acyclic graph (DAG), existing methods often lack efficiency, making them unsuitable for online applications. In this paper, we propose \sysname{}, an efficient multi-agent RL-based approach for incremental DAG learning. \sysname{} uses a DAG generation policy that maps a continuous real-valued space to the DAG space as an intra-batch strategy, then incorporates two RL agents—state-specific and state-invariant—to uncover causal relationships and integrates these agents into an incremental learning framework. Furthermore, the framework leverages a factored action space to enhance parallelization efficiency. Extensive experiments on synthetic and real datasets demonstrate that \sysname{} outperforms state-of-the-art methods in terms of both efficiency and effectiveness. 
\end{abstract}


\section{Introduction}

Discovering and understanding the causal mechanisms behind natural phenomena is crucial in many scientific fields~\cite{rl-bic}. Consequently, methods for discovering causality from observational data have gained significant attention. Understanding these causal relationships helps predict the outcomes of interventions and hypothetical scenarios, which is highly valuable in areas like econometrics~\cite{wold1954causality}. Due to the constantly changing nature of these fields, there is an increasing need to develop and use causal discovery methods in online settings to improve real-time decision-making and adaptability.

\begin{figure}[t]
    \centering
    \includegraphics[width=0.45\textwidth]{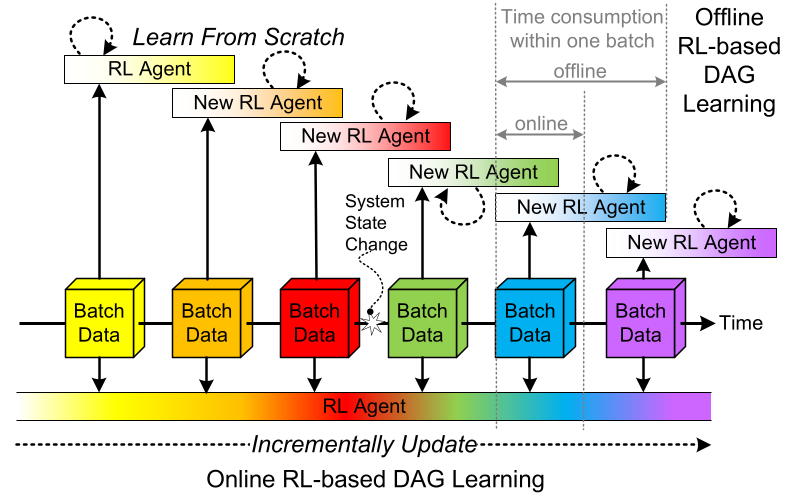}

     \caption{Comparison of the learning processes of offline (top) and online (bottom) RL-based DAG learning methods in an online data stream. The gradient colors on the RL agent represent its learning progress, with white indicating the initial state. When the color aligns with a data batch, it signifies that the agent has learned the current causal mechanism. Instead of learning from scratch, online DAG learning needs to incrementally and efficiently adapt to continuously arriving data batches and non-stationary data distributions.}   
    \label{fig:online_intro}

\end{figure}

Identifying causal structures involves finding a Directed Acyclic Graph (DAG) $\mathcal{\hat{G}}$ that minimizes a score function $\mathcal{S}$ based on observed data \textbf{X}:
\begin{equation}
\label{eq:score_based_obj}
    \mathop{min}_\mathcal{\hat{G}} \mathcal{S}(\mathcal{\hat{G}}, \textbf{X}),\  \text{s.t. } \mathcal{\hat{G}} \in \text{DAGs}.
\end{equation}

However, this process is NP-hard~\cite{corl} due to the super-exponential growth of the DAG space with the number of nodes~\cite{robinson1977counting} and the acyclicity constraint. Numerous methods have been proposed~\cite{notears, golem, rl-bic,li2025solverllm, corl} to tackle this problem, with continuous optimization methods recently receiving significant attention. Pioneered by NOTEARS~\cite{notears}, these methods transform the problem into a continuous optimization task by introducing a smooth feature to ensure acyclicity. However, they still tend to get stuck in local optima.

Reinforcement learning (RL) has become a promising approach for DAG learning due to its effective search strategies and explainable rewards~\cite{rl-bic,corl,zhao2025uncertainty,rcl-og}. RL-BIC~\cite{rl-bic} explores the full graph space and penalizes cycles via rewards but can't fully ensure acyclicity and is resource-intensive~\cite{corl}. Ordering-based methods~\cite{rcl-og} operate in the ordering space to bypass acyclicity issues but rely on sequential decisions, limiting parallelization. These limitations hinder RL-based methods from scaling to real-world problems. 
Recent work ALIAS~\cite{alias} maps a real-valued vector to the DAG space and performs causal discovery without acyclicity constraints, enabling efficient RL.

Moreover, most existing RL-based methods are designed for offline settings, neglecting the more practical scenario of incremental DAG learning in online environments. Online DAG learning is particularly valuable for handling continuous data streams generated by modern applications, as it allows models to be updated incrementally with new data. This not only optimizes resource utilization but also enables real-time analysis and decision-making.  
When applied to online settings, RL-based methods encounter additional challenges due to stringent real-time requirements, as illustrated in Figure~\ref{fig:online_intro}. First, online causal discovery demands that RL algorithms process incoming data batches efficiently, placing strict constraints on intra-batch performance. Second, unlike offline approaches that can be retrained from scratch, online methods must incrementally refine the model by integrating newly acquired information. This allows them to rapidly adapt to evolving system states and capture causal mechanisms under shifting data distributions, ensuring more effective and timely decision-making. 

To address these challenges, in this paper, we propose \sysname{}, an efficient \underline{m}ulti-\underline{a}gent \underline{r}einforcement \underline{l}earn\underline{in}g framework for incremental DAG learning. 
We first develop an efficient intra-batch DAG learning method that maps from a continuous real-valued space to the DAG space without enforcing an acyclicity constraint. Building on this, we develop two RL agents to incrementally learn and disentangle state-invariant and state-specific causation from non-stationary online data across different batches, efficiently discovering the DAG continuously.
Furthermore, we factor the action space to enable parallel DAG learning for efficiency improvement. Extensive experiments on both synthetic and real datasets validate the effectiveness and efficiency of \sysname{} in incremental DAG learning.

\section{Related Work}

\noindent\textbf{Traditional DAG Learning Methods.} Existing DAG learning methods fall into four categories: \textbf{(i) Constraint-based methods}, like the PC algorithm~\cite{pc}, use conditional independence (CI) tests to recover DAG structures but depend heavily on CI test accuracy, making them unreliable when conflicts arise. \textbf{(ii) Score-based methods} evaluate DAGs using scoring functions (\textit{e.g.}, BIC~\cite{BIC}), but their combinatorial search complexity limits scalability. \textbf{(iii) Continuous optimization methods} reformulate DAG learning as a smooth optimization problem, as in NOTEARS~\cite{notears}, but often struggle to escape local optima~\cite{dag-gnn}. \textbf{(iv) Sampling-based methods} estimate DAG posteriors but are computationally expensive, even with recent differentiable sampling techniques~\cite{charpentier2022differentiable}. Traditional methods focus on offline settings and face efficiency challenges. In contrast, our approach, \sysname{}, employs multi-agent reinforcement learning to efficiently search for global optima and adapt to online settings.

\noindent\textbf{Reinforced DAG Learning.} In recent years, RL has become a promising approach for combinatorial optimization, offering an intuitive search process and interpretable rewards to overcome local heuristic limitations. RL-BIC~\cite{rl-bic} trains an RL agent to find high-reward DAGs with implicit acyclicity penalties but searches the entire directed graph space, making it inefficient and unable to strictly guarantee acyclicity. Ordering-based methods like CORL~\cite{corl} and RCL-OG~\cite{rcl-og} reduce the search space by framing variable ordering as an MDP, mapping orderings to fully-connected DAGs~\cite{teyssier2012ordering}, and estimating DAGs via variable selection. While these methods avoid directly dealing with acyclicity, their sequential nature makes them unsuitable for parallelization and less efficient in online settings. Recent work ALIAS~\cite{alias} projects a real-valued vector into the DAG space, allowing causal discovery without enforcing acyclicity constraints, which facilitates efficient RL, making continuous and efficient DAG discovery possible.
However, these methods focus solely on the offline setting, limiting their real-world applications. In contrast, our approach enables efficient and continuous DAG discovery through multi-agent RL.

\section{Preliminaries}

\noindent \textbf{Structural Equation Model (SEM).} Let $\mathcal{G}=\{\mathcal{V},\mathcal{E}\}$ denote a DAG, where each node $v_i \in \mathcal{V}=\{v_1,...,v_d\}$ is associated with a random variable $X_i \in \mathcal{X}=\{X_1,...,X_d\}$. Each directed edge $(v_i,v_j)\in\mathcal{E}=\{(v_i,v_j)|i,j=1,...,d \ \text{and}\ i\ne j\}$ indicates that $X_i$ is a direct cause of $X_j$. The DAG $\mathcal{G}$ can be equally represented by a binary adjacency matrix $\textbf{A} \in\{0,1\}^{d\times d}$, where the $(i,j)$-th entry is 1 if $(v_i,v_j) \in \mathcal{E}$ and 0 otherwise. The joint distribution associated with $\mathcal{G}$ can be decomposed into $P(X_1,...,X_d)=\prod_{i=1}^dP(X_i|\text{Pa}(X_i))$, where $\text{Pa}(X_i) = \{X_k \mid (v_k,v_i) \in \mathcal{E}\}$ is the set of parents of $X_i$ in $\mathcal{G}$. We assume that the data generation process conforms to a SEM with additive noise:
\begin{equation}
\label{eq:SEM}
    X_i=f_i(\text{Pa}(X_i))+\eta_i, i=1,...,d,
\end{equation}
where $f_i(\cdot)$ represents the causal relationship between $X_i$ and its parents $\text{Pa}(X_i)$, and the additive noise $\eta_i$ is assumed to be jointly independent. We also assume \textit{causal minimality}, meaning that each $f_i(\cdot)$ is not constant with respect to any of its arguments~\cite{peters2014causal}.

\begin{figure*}[t]
    \centering
    \includegraphics[width=0.95\textwidth]{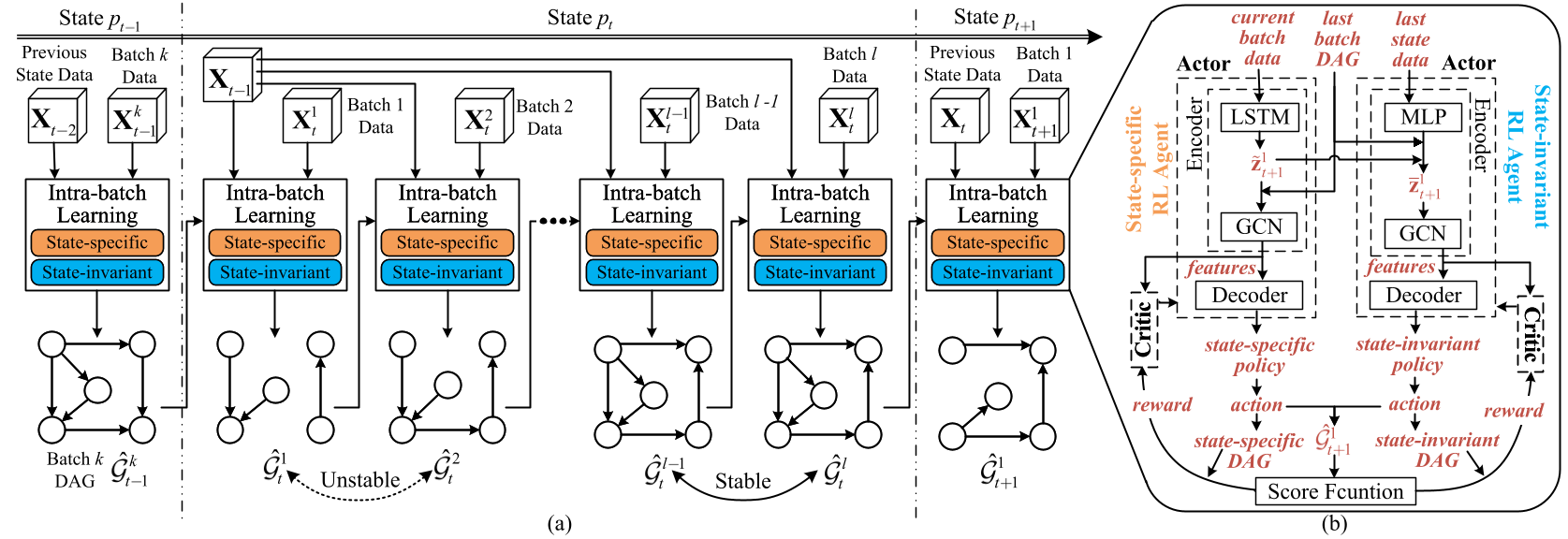} 
    \caption{(a) The pipeline of \sysname{} across three consecutive system states. For each batch, \sysname{} learns the DAG using the intra-batch single-step RL algorithm, which includes state-specific and state-invariant RL agents optimizing their policies through an actor-critic approach. Detailed network architecture and variables are explained in Section~\ref{sec:incremental_learning}. \sysname{} facilitates efficient incremental DAG learning by disentangling state-specific and state-invariant causal relationships.}
    \label{fig:overall}
\end{figure*}

\noindent \textbf{Non-Stationarity of Online Data.} Real-world online data streams frequently exhibit non-stationarity~\cite{shao2024supervised}, with the causal relationships among random variables being dynamic and subject to change over time. We assume that such changes are sufficient to induce partial changes in the underlying causal structure, leading to transitions in system state\footnote{We distinguish between ``system state'' and ``state'' as different concepts. The former refers to the underlying causal mechanisms of a system, while the latter is a concept used in RL.}. In this context, although the new system state introduces causal relationships that are specifically dependent on it (\textit{i.e.,} system \textbf{state-specific} causation), some inherent system causal relationships remain unchanged over time, reflecting the \textbf{state-invariant} aspects of the system's dynamics.

\noindent \textbf{Problem Statement.} Given a dataset $\mathcal{D}=\{\textbf{X}_t\}_{t=1}^m$ comprising $m$ sequentially continuous sets of observations $\textbf{X}_t\in\mathbb{R}^{n_t\times d}$. Each $\textbf{X}_t$ corresponds to a system state $p_t$, with $n_t$ observations, and is associated with a DAG $\mathcal{G}_t$. In an online setting, data for each system state $p_t$ arrives in batches of size $b$, denoted by $\textbf{X}_t = [(\textbf{X}_t^1)^\top,...,(\textbf{X}_t^L)^\top]^\top$, where $\textbf{X}_t^l\in\mathbb{R}^{b\times d}, l=1,...,L$, represents the $l$-th batch of $\textbf{X}_t$ and is associated with the DAG $\mathcal{G}_t^l$, which captures the causal mechanisms of the current batch data. Our goal is to \textbf{efficiently} perform DAG learning on each batch and \textbf{effectively} learn the estimated DAG $\mathcal{\hat{G}}_t$ for each system state $p_t$, aiming to achieve the best average score across all system states:
\begin{equation}
\label{eq:our_online_obj}
    \mathop{min}_{\mathcal{\hat{G}}_1,...,\mathcal{\hat{G}}_m} \frac{1}{m}\sum_{t=1}^m\mathcal{S}(\mathcal{\hat{G}}_t, \textbf{X}_t),\  \text{s.t. } \mathcal{\hat{G}}_t \in \text{DAGs}, \quad \mathcal{\hat{G}}_t \ne \mathcal{\hat{G}}_{t+1}, \forall t.
\end{equation}

System state transition detection is beyond the scope of this work. Here, we use the Multivariate Singular Spectrum Analysis model~\cite{alanqary2021change}, an effective method for online change point detection, to identify state transition points in the online DAG learning process. 


\section{Methodology}

We propose \sysname{}, a multi-agent reinforcement learning framework designed for incremental DAG learning. As illustrated in Figure~\ref{fig:overall}, \sysname{} involves two modules: (i) Intra-batch Reinforced DAG Learning: We develop an intra-batch DAG generation method using two matrices derived solely from a real-valued vector, thereby mapping a continuous real space to the DAG space; 
(ii) Incremental Multi-agent Reinforced DAG Learning: This module leverages multi-agent reinforcement learning with state-specific and state-invariant agents to learn and disentangle state-invariant and state-specific causation across different batches. The approach is integrated into an incremental learning framework to efficiently uncover causal relationships. In addition, we explore the potential for parallel computation within this framework.

\subsection{Intra-batch Reinforced DAG Learning}
\label{sec:intra-batch_learning}

In this subsection, inspired by~\cite{massidda2023constraint,alias}, we learn a DAG in a single batch through the score-based causal discovery method based on one-step reinforcement learning.

Sampling DAGs from a parameterized distribution is crucial for exploring the DAG space effectively. To ensure acyclicity, there is a well-established decomposition technique~\cite{charpentier2022differentiable} that decomposes a DAG into two binary matrices:
\begin{equation}
\label{eq:DAG_decompose}
    \textbf{A} = \textbf{P}^\top\textbf{U}\textbf{P},
\end{equation}
where $\textbf{A}\in\{0,1\}^{d\times d}$ is the adjacency matrix of a DAG $\mathcal{G}$, $\textbf{P}\in\{0,1\}^{d\times d}$ is a permutation matrix, and $\textbf{U}\in\{0,1\}^{d\times d}$ is a strictly upper-triangular matrix. The matrix $\textbf{U}$ represents the adjacency matrix of a graph that \textbf{ensures acyclicity}, with all directed edges $(v_i,v_j)$ satisfying $i < j$. This captures all subsets of a specific fully-connected (FC) DAG corresponding to the ``initial'' ordering of nodes, where node $v_i \in \mathcal{V}$ is in the $i$-th position. The permutation matrix $\textbf{P}$ changes the order of nodes in this initial ordering, resulting in a graph with the same topological structure. Consequently, $\textbf{P}$ represents all subsets of the FC DAG corresponding to the altered ordering, allowing Eq.~\ref{eq:DAG_decompose} to cover the entire DAG space.
Based on this intuition, an arbitrary DAG $\textnormal{\textbf{A}}\in\{0,1\}^{d\times d}$ can be obtained from a FC DAG \textbf{H} and a binary mask matrix \textbf{S}, as shown below:

\begin{equation}
   \label{eq:dag_sampling}
    \textnormal{\textbf{A}} = \textnormal{\textbf{H}} \odot \textnormal{\textbf{S}},
    \end{equation}
where $\odot$ is the Hadamard product operator.
Instead of time-consuming ordering-based methods to obtain a fully-connected matrix, we derives \textbf{H} from a single real-valued vector \textbf{h}:
\begin{equation}
   \label{eq:fc_matrix}
    H_{ij}=
    \begin{cases} 
    1,  & \text{if }\ h_i > h_j, \\
    0, & \text{otherwise}.
    \end{cases}
    \end{equation}

Since \( \textbf{S} \) can be easily obtained by filtering a real-valued matrix of the same shape with a simple threshold to produce a binary matrix, for any given real-valued vector \( \textbf{a} \) of dimension \( d(d+1) \), we can generate a matrix \( \textbf{H} \) from its first \( d \) dimensions and a matrix \( \textbf{S} \) from the subsequent \( d^2 \) dimensions, thereby obtaining an arbitrary \( \textbf{A} \).

Based on the above idea, a single-step RL algorithm employs a stochastic policy \( \pi \) to learn an action \( a \), which can be directly used to search the DAG space. The policy \( \pi \) selects a continuous action \( a \) from the real-valued space, which in turn determines the DAG of \( d \) nodes. After that, the reward \( \mathcal{R} \) is calculated using a score function \( \mathcal{S} \): $\mathcal{R}(\textbf{a},\textbf{X})=-\mathcal{S}(g(\textbf{a}),\textbf{X})$. Specifically, we use the Bayesian Information Criterion (BIC) score~\cite{BIC} to enable straightforward comparisons with other RL-based methods.

\subsection{Incremental Multi-Agent Reinforced DAG Learning}
\label{sec:incremental_learning}

Incremental learning enables a model to update itself as new data arrives~\cite{masana2022class}, eliminating the need for retraining from scratch. To tackle the problem outlined in Eq.~\ref{eq:our_online_obj}, we propose a novel incremental learning framework based on multi-agent RL. Each agent builds upon the one-step reinforced DAG learning module described in Section~\ref{sec:intra-batch_learning} and illustrated in Figure~\ref{fig:overall} (b). Specifically, we employ a state-invariant RL agent to incrementally learn causal relationships that remain consistent across different system states, and a state-specific RL agent to swiftly identify causal relationships unique to the current system state. This disentanglement mechanism enables \sysname{} to efficiently capture state-specific causal mechanisms by leveraging the incrementally updated state-invariant information as acquired knowledge when encountering new data distributions, thus facilitating effective inter-batch incremental DAG learning. 

\noindent\textbf{State-specific RL Agent.} Suppose the incoming data represents the $l$-th batch for system state $p_t$. The state-specific RL agent's objective is to learn the new causal relationships introduced by the data batch $\textbf{X}_t^l$. To capture information changes across different batches, the encoding component utilizes both $\textbf{X}_t^l$ and the previous hidden state as inputs to a long short-term memory network (LSTM)~\cite{hochreiter1997long}, producing the embedding $\tilde{\textbf{z}}_t^l$ for current batch. This embedding is then combined with the DAG from the previous batch $\mathcal{G}_t^{l-1}$, to create an attributed graph. This graph, which incorporates prior structural knowledge, is subsequently encoded using a graph convolutional network (GCN) \cite{kipf2016semi}. 

Then, a decoder is used to learn a state-specific policy $\tilde{\pi}_t^l$ that samples an action $\tilde{\textbf{a}}_t^l$ to generate the state-specific DAG $\tilde{\mathcal{G}}_t^l$. This action is then combined with the $\bar{\textbf{a}}_t^l$, which is sampled from the state-invariant policy $\bar{\pi}_t^l$, to produce the fusion action $\hat{\textbf{a}}_t^l = \beta\tilde{\textbf{a}}_t^l+(1-\beta)\bar{\textbf{a}}_t^l$, where $\beta\in[0,1]$ is used to balance the importance of state-specific information and state-invariant information. And then, the complete DAG $\mathcal{G}_t^l$ for $\textbf{X}_t^l$ is obtained based on Section 4.1 via $\hat{\textbf{a}}_t^l$. By default, we set the $\beta$ to be $0.5$. 

To ensure accurate discovery of state-specific information, we introduce a decoupling term in the reward function to encourage the estimated state-specific DAG $\tilde{\mathcal{G}}_t^l$ to be as distinct as possible from both the previous batch's state-invariant DAG $\bar{\mathcal{G}}_t^{l-1}$ and the DAG from the previous state  $\mathcal{G}_{t-1}$. This term is defined as follows:
\begin{equation}
   \label{eq:specific_term}
    \mathcal{L}_{\tilde{\mathcal{G}}_t^l}=\left(\|\tilde{\textbf{A}}_t^l-\complement{\bar{\textbf{A}}_t^{l-1}}\|^2 + \|\tilde{\textbf{A}}_t^l-\complement{\textbf{A}_{t-1}}\|^2\right)/d,
\end{equation}
where $d$ is the number of nodes in \textbf{A} and $\complement{\textbf{A}}$ denotes the complement of the adjacency matrix \textbf{A}, which involves converting 0s in \textbf{A} to 1s and 1s to 0s. The current reward is defined as $\tilde{\mathcal{R}}_t^l=-\mathcal{S}(\textbf{A}_t^l,\textbf{X}_t^l) + \lambda_1 \mathcal{L}_{\tilde{\mathcal{G}}_t^l}$, where $\lambda_1$ represents the weight balancing the decoupling term and the BIC score. Since state-specific information is highly dependent on the system state, the state-specific RL agent is reinitialized at the beginning of each new system state.

\noindent\textbf{State-Invariant RL Agent.} The state-invariant RL agent aims to learn the causal relationships that remain consistent across multiple system states. The encoding process begins by using a fully connected layer to transform the previous state data $\textbf{X}_{t-1}$ into embedding $\bar{\textbf{z}}_{t-1}$. Given that state-invariant causal relationships are influenced by both $\textbf{X}_{t-1}$ and $\textbf{X}_t^l$, we concatenate $\bar{\textbf{z}}_{t-1}$ and $\tilde{\textbf{z}}_t^l$ to form $\bar{\textbf{z}}_t^l$. The subsequent steps are similar to those in the state-specific RL agent: after encoding the attributed graph formed by $\mathcal{G}_t^{l-1}$ and $\bar{\textbf{z}}_t^l$ using a GCN, the state-invariant policy $\bar{\pi}_t^l$ is learned through a decoder to generate the state-invariant DAG $\bar{\mathcal{G}}_t^l$. This DAG is then combined with $\tilde{\mathcal{G}}_t^l$ to produce $\mathcal{G}_t^l$. Additionally, a decoupling term is introduced to ensure that the estimated state-invariant DAG $\bar{\mathcal{G}}_t^l$ remains as dissimilar as possible to the previous batch's state-specific $\tilde{\mathcal{G}}_t^{l-1}$ while staying similar to $\mathcal{G}_{t-1}$. This is defined as:
\begin{equation}
   \label{eq:invariant_term}
    \mathcal{L}_{\bar{\mathcal{G}}_t^l}=\left(\|\bar{\textbf{A}}_t^l-\complement{\tilde{\textbf{A}}_t^{l-1}}\|^2 + \|\bar{\textbf{A}}_t^l-\textbf{A}_{t-1}\|^2\right)/d.
\end{equation}
The reward is then defined as $\bar{\mathcal{R}}_t^l=-\mathcal{S}(\textbf{A}_t^l,\textbf{X}_t^l) + \lambda_2 \mathcal{L}_{\bar{\mathcal{G}}_t^l}$, where $\lambda_2$ is a weight coefficient. Since state-invariant information remains constant over time, the state-invariant RL agent is continuously updated throughout learning.

\noindent\textbf{Optimization.} Both agents are trained using the Adam optimizer~\cite{kinga2015method}. We introduce a baseline for more stable training~\cite{sutton2018reinforcement}, so that each agent's objective is to minimize the temporal difference (TD) error between the critic’s predicted rewards $\hat{\mathcal{R}}$ plus a baseline $\mathcal{B}$ and its actual rewards $\mathcal{R}$. The baseline $\mathcal{B}$ is updated according to the formula:
$\mathcal{B} = \gamma \cdot \mathcal{B} + (1 - \gamma) \cdot \bar{R},
$ where $\gamma$ is the discount factor and $\bar{R}$ denotes the mean of the rewards $\mathcal{R}$.
The policy gradient is given by $\nabla J(\boldsymbol{\psi})=\mathbb{E}_{\pi({\boldsymbol{\psi}})}\{\nabla_{\boldsymbol{\psi}}\log{\pi({\boldsymbol{\psi}})[\mathcal{R}-(b+\hat{\mathcal{R}})]}\}$.

\noindent\textbf{Model Convergence within State.}
The estimated DAG is expected to gradually converge as successive data batches are processed. To avoid wasting computational resources, we define the similarity between the estimated DAGs of two consecutive batches within the same system state $p_t$ using the Jensen-Shannon (JS) divergence~\cite{fuglede2004jensen}:
\begin{equation}
    \label{eq:graph_similarity}
     \xi = 1 - \text{JS}(P_{\mathcal{G}}(\mathcal{G}_t^{l-1})||P(\mathcal{G}_t^l)),
\end{equation}
where $P_{\mathcal{G}}(\cdot)$ denotes the edge distribution of graph. A larger $\xi$ indicates a greater similarity between the two graphs. When $\xi$ exceeds a certain threshold, we consider the current estimated DAG to have stabilized and will terminate the learning process for the current system state early, until a new system state arrives.


    


\subsection{Factored Action Space for Parallelization}
\label{sec:multi-agent}

Our method operates in a single step, in contrast to the multi-step decision processes of ordering-based methods. Each position in our action vector \textbf{a} has a specific role in forming the final DAG (\textit{e.g.}, the first $d$ elements of \textbf{a} are used as \textbf{h} to obtain \textbf{H}). Thus, the action space can be decomposed into multiple subspaces, transforming our problem into one involving a factored action space~\cite{tang2022leveraging}, which can then be parallelized across multiple processing units. Each unit explores a subspace, and the combination of these subspaces forms the complete action space, significantly enhancing efficiency when applied to online applications. 
We refer to this variant of factored action space as \sysnamevariant{}.

\section{Experiments}

\begin{figure*}[ht]

    \centering
    \begin{subfigure}[b]{1\textwidth}
        \centering
        \includegraphics[width=0.95\textwidth]{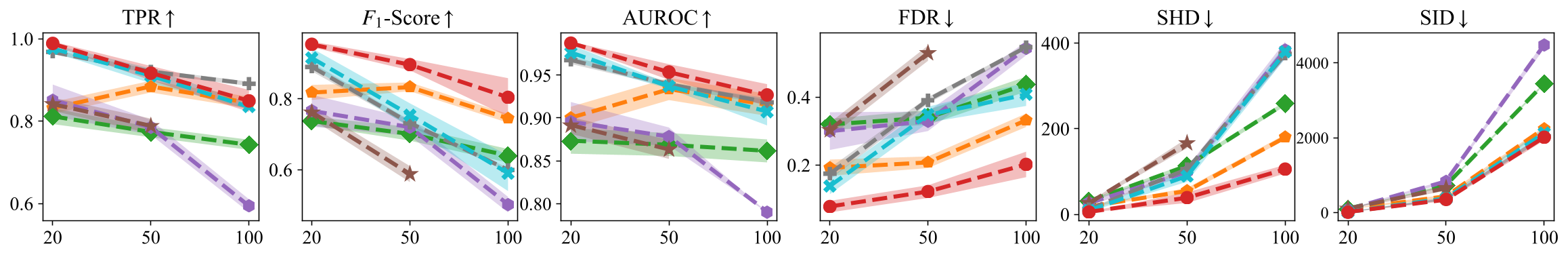}
    \caption{DAG learning performance across different DAG scales (with $d$ as x-axis).}
    \label{fig:synthetic_linear_scale}
    \end{subfigure}
    \hfill
    \begin{subfigure}[b]{1\textwidth}
        \centering
        \includegraphics[width=0.95\textwidth]{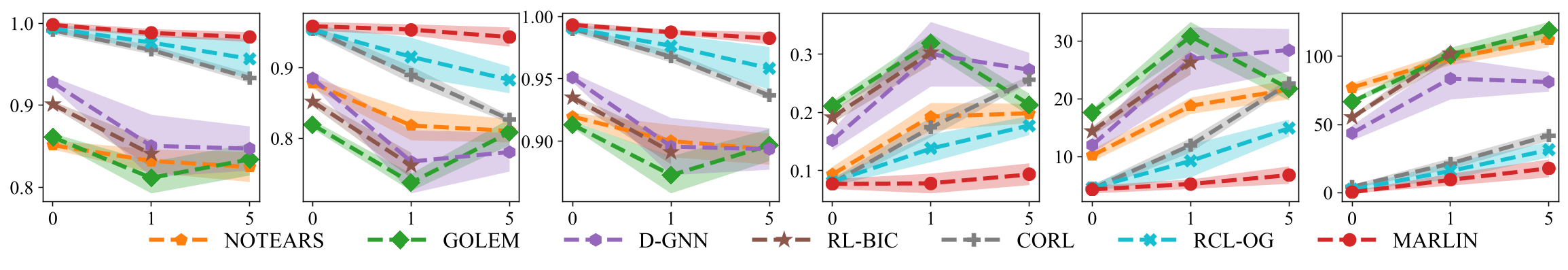}
            \caption{DAG learning performance across different transition noise rates (with $e$ as x-axis).}
            \label{fig:synthetic_linear_noise}
    \end{subfigure}
    \caption{Average performance of DAG learning across all states on synthetic Linear-Gaussian datasets, varying (a) DAG scale ($d=\{20,50,100\}$) and (b) transition noise rate ($e=\{0,1,5\}$) for \sysname{} and other baselines. The shaded area represents the standard deviation; $\uparrow$ indicates that higher values are better, while $\downarrow$ indicates that lower values are better.}

    \label{fig:synthetic_linear_result}
\end{figure*}

\subsection{Experimental Setup}

\subsubsection{Datasets} We conduct extensive experiments with various synthetic datasets of differing scales and generation methods (see Section \ref{sec:synthetic_gen} for details) to evaluate the algorithm's capacity for incremental learning of DAGs. Furthermore, to evaluate \sysname{}'s effectiveness in real-world scenarios, we apply it to causal discovery-based root cause analysis (RCA) tasks~\cite{zheng2024mulan,wang2023interdependent,wang2023incremental,wang2023hierarchical,zheng2024online} using three time series datasets from real systems: \textbf{(i) OnlineBoutique (OB)}~\cite{yu2023nezha} is a microservice system for e-commerce composed of $10$ microservices, which experienced $18$ system faults during the data collection period. \textbf{(ii) Secure Water Treatment (SWaT)}~\cite{mathur2016swat,zheng2024lemma} is a scaled-down model of a real industrial water treatment plant, equipped with $51$ sensors and actuators. The dataset includes $3$ hours of SWaT operation under normal conditions and $1$ hour during which $6$ attacks were executed. \textbf{(iii) Water Distribution (WADI)}~\cite{ahmed2017wadi,zheng2024lemma} consists of data collected from a water distribution testbed with $123$ sensors and actuators over $16$ days of continuous operation, including $14$ days under normal conditions and $2$ days with $15$ fault cases.

\subsubsection{Synthetic Data Generation}
\label{sec:synthetic_gen}

We introduce a synthetic data generation strategy tailored to our online setting. Starting from a complete Erdős–Rényi (ER) DAG with $d$ nodes, we iteratively construct incomplete DAGs by randomly deleting edges and injecting $e\%$ noisy edges while preserving acyclicity. Observations are then generated following RL-BIC \cite{rl-bic}, ordered from the most incomplete to the complete DAG, yielding datasets that capture system state transitions for validating incremental DAG learning.

Here, we focus primarily on three factors: the scale of the DAG (number of nodes $d$), the observations generation procedure (the type of regression methods for the causal mechanisms and noise), and the rate of noise added during the system state transition process (referred to as the transition noise rate $e\%$). Based on variations in these factors, we generated multiple synthetic datasets.

\subsubsection{Baselines} We compare \sysname{} with seven DAG learning algorithms: the classical constraint-based algorithm PC~\cite{pc}, continuous optimization algorithms NOTEARS~\cite{notears}, GOLEM~\cite{golem}, and DAG-GNN~\cite{dag-gnn}, as well as RL-based algorithms RL-BIC~\cite{rl-bic}, CORL~\cite{corl}, and RCL-OG~\cite{rcl-og}. Since these algorithms are designed to learn DAGs in offline settings, they lack incremental learning mechanisms, which prevents them from inheriting causations from previous data batches. Consequently, each new data batch requires learning from scratch to adapt to an online setting.

\subsubsection{Evaluation Metrics} We assess the performance using six common metrics: True Positive Rate (TPR), $F_1$-score, Area Under the Receiver Operating Characteristic Curve (AUROC), False Discovery Rate (FDR), Structural Hamming Distance (SHD), and Structural Intervention Distance (SID)~\cite{SID}. For the estimated graph, higher values of TPR, $F_1$-score, and AUROC are preferred, whereas lower values of FDR, SHD, and SID are desirable. We use the first state as historical offline data to train the initial model and then comprehensively evaluated the estimated DAGs learned by the algorithm on all subsequent states. We average the results across all states to evaluate the algorithm's performance over the entire dataset. Additionally, the average running time per batch (ATB) is calculated to measure the algorithm's efficiency.

\begin{table}[t]
    \scriptsize
    \centering
    \setlength\tabcolsep{0.5pt}

    \begin{tabular}{l|lcccccc}
        \toprule
         \multicolumn{2}{c}{} & TPR$\uparrow$ & FDR$\downarrow$ & SHD$\downarrow$ & AUROC$\uparrow$ & SID$\downarrow$ & ATB$\downarrow$\\
        \midrule
       \multirow{8}{*}{QR\ } & \ PC & 0.30\scriptsize{(0.01)} & 0.65\scriptsize{(0.02)} & 83.2\scriptsize{(1.3)} & 0.58\scriptsize{(0.00)} & 330.0\scriptsize{(6.8)}  & $-$ \\
        & \ NOTEARS & 0.28\scriptsize{(0.01)} & 0.15\scriptsize{(0.03)} & 78.0\scriptsize{(1.9)} & 0.63\scriptsize{(0.00)} & 270.0\scriptsize{(6.5)}  & $-$ \\
        & \ GOLEM & 0.36\scriptsize{(0.01)} & 0.35\scriptsize{(0.02)} & 78.2\scriptsize{(1.4)} & 0.65\scriptsize{(0.01)} & 245.0\scriptsize{(8.6)}  &  $-$\\
        & \ D-GNN & 0.31\scriptsize{(0.00)} & 0.38\scriptsize{(0.05)} & 86.5\scriptsize{(3.1)} & 0.63\scriptsize{(0.01)} & 264.0\scriptsize{(5.9)}  &  $-$\\
        & \ RL-BIC & 0.84\scriptsize{(0.02)} & 0.30\scriptsize{(0.02)} & 26.3\scriptsize{(2.1)} & 0.89\scriptsize{(0.01)} & 102.0\scriptsize{(6.1)}  &  330\scriptsize{(60)}\\
        & \ CORL & 0.88\scriptsize{(0.01)} & 0.25\scriptsize{(0.01)} & 21.1\scriptsize{(0.7)} & 0.92\scriptsize{(0.01)} & 78.1\scriptsize{(11.1)}  &  416\scriptsize{(38)}\\
        & \ RCL-OG & 0.90\scriptsize{(0.03)} & 0.18\scriptsize{(0.07)} & 15.6\scriptsize{(6.2)} & 0.94\scriptsize{(0.02)} & 68.9\scriptsize{(18.1)}  & 266\scriptsize{(42)} \\
        \cmidrule(lr){2-8}
        & \textbf{\ \sysname{}} & \textbf{0.94\scriptsize{(0.01)}} & \textbf{0.08\scriptsize{(0.01)}} & \textbf{7.0\scriptsize{(0.7)}} & \textbf{0.96\scriptsize{(0.00)}} & \textbf{49.6\scriptsize{(7.1)}}  & 81\scriptsize{(9)} \\
        & \ \sysnamevariant{} & 0.90\scriptsize{(0.01)} & 0.15\scriptsize{(0.01)} & 14.2\scriptsize{(0.4)} & 0.92\scriptsize{(0.00)} & 65.1\scriptsize{(5.6)}  & \textbf{32\scriptsize{(3)}} \\
       \midrule
       \multirow{8}{*}{GP\ } & \ PC & 0.19\scriptsize{(0.01)} & 0.71\scriptsize{(0.00)} & 74.8\scriptsize{(1.2)} & 0.55\scriptsize{(0.01)} & 315.2\scriptsize{(3.2)}  & $-$ \\
        & \ NOTEARS & 0.29\scriptsize{(0.00)} & 0.58\scriptsize{(0.01)} & 63.4\scriptsize{(1.7)} & 0.61\scriptsize{(0.01)} & 265.0\scriptsize{(4.2)}  & $-$ \\
        & \ GOLEM & 0.36\scriptsize{(0.01)} & 0.51\scriptsize{(0.01)} & 43.4\scriptsize{(0.9)} & 0.65\scriptsize{(0.00)} & 241.8\scriptsize{(4.8)}  &  $-$\\
        & \ D-GNN & 0.25\scriptsize{(0.01)} & 0.62\scriptsize{(0.06)} & 70.0\scriptsize{(2.8)} & 0.58\scriptsize{(0.01)} & 295.4\scriptsize{(6.4)}  &  $-$\\
        & \ RL-BIC & 0.80\scriptsize{(0.01)} & 0.35\scriptsize{(0.04)} & 31.3\scriptsize{(2.2)} & 0.86\scriptsize{(0.02)} & 159.8\scriptsize{(8.5)}  &  415\scriptsize{(88)}\\
        & \ CORL & 0.86\scriptsize{(0.01)} & 0.27\scriptsize{(0.01)} & 26.3\scriptsize{(3.2)} & 0.88\scriptsize{(0.01)} & 105.1\scriptsize{(11.6)}  & 455\scriptsize{(34)} \\
        & \ RCL-OG & 0.87\scriptsize{(0.01)} & 0.23\scriptsize{(0.04)} & 20.4\scriptsize{(2.2)} & 0.92\scriptsize{(0.02)} & 98.9\scriptsize{(15.4)} & 293\scriptsize{(30)} \\
        \cmidrule(lr){2-8}
        & \ \textbf{\sysname{}} & \textbf{0.92\scriptsize{(0.01)}} & \textbf{0.15\scriptsize{(0.02)}} & \textbf{13.2\scriptsize{(1.6)}} & \textbf{0.95\scriptsize{(0.02)}} & \textbf{78.9\scriptsize{(10.3)}}  & 85\scriptsize{(6)} \\
        & \ \sysnamevariant{} & 0.87\scriptsize{(0.01)} & 0.20\scriptsize{(0.01)} & 18.8\scriptsize{(0.9)} & 0.91\scriptsize{(0.01)} & 102.8\scriptsize{(9.3)}  & \textbf{33\scriptsize{(4)}} \\
    
        \bottomrule
    \end{tabular}
    \caption{DAG learning performance on synthetic QR and GP datasets with standard deviations reported in parentheses. A dash ($-$) for ATB indicates methods with notably poor performance. Bold indicates the best performance; $\uparrow$ indicates higher is better, and $\downarrow$ indicates lower is better.}
    \label{tab:synthetic_other_result}
\end{table}

\subsection{Performance Evaluation}

\subsubsection{Linear Model with Gaussian Noise}
\label{sec:result_syn_lg}

We first evaluate the effectiveness of \sysname{} using Linear-Gaussian (LG) synthetic datasets. These datasets are generated by applying a linear model with Gaussian noise, varying both the DAG scale ($d=\{20,50,100\}$) and the transition noise rate ($e=\{0,1,5\}$).
The PC algorithm's results are excluded from the main discussion due to its suboptimal performance on dense graphs.

Figure~\ref{fig:synthetic_linear_result} reports results on these datasets, yielding four insights: (1) \sysname{} consistently surpasses nearly all baselines across metrics, highlighting its strong DAG learning and robustness to state transitions. (2) CORL achieves higher TPR than \sysname{} at $d=100$, but due to overly dense graphs with spurious edges, reflected in poor FDR and SHD—a pattern also seen in RCL-OG. Both methods degrade notably under increasing transition noise. (3) RL-BIC, though RL-based, struggles with incremental DAG learning, worsening as graph size grows and failing at $d=100$. (4) NOTEARS shows some promise in dynamic settings but still trails \sysname{}, while GOLEM and DAG-GNN perform poorly, with DAG-GNN especially unstable as scale and complexity increase.

\subsubsection{Non-Gaussian or Nonlinear Models}
To further assess the robustness of the algorithms under non-Gaussian noise and nonlinear models, we employ three distinct generation procedures, maintaining $d=20$ and $e=1$ constant: (i) a linear model with exponential noise (LE), (ii) a nonlinear model with quadratic regression (QR), and (iii) a nonlinear model utilizing Gaussian processes (GP). Performance and runtime efficiency of \sysname{}, \sysnamevariant{}, and the baseline methods on the remaining two synthetic datasets are presented in Table~\ref{tab:synthetic_other_result}. 


We make four key observations: First, \sysname{} surpasses all baselines in both efficiency and accuracy across all datasets, showing strong effectiveness on nonlinear data. Second, \sysnamevariant{} learns DAGs close in quality to \sysname{} while offering greater stability and faster runtime, indicating that parallel computation enables real-time use with minimal performance loss. Its slight accuracy drop likely arises from the decomposition and reconstruction of the action space, which limits holistic causal modeling but remains an acceptable efficiency trade-off. Third, non-RL methods perform poorly because they struggle with nonlinear causal structures. Finally, although other RL-based methods handle nonlinear models well, their high computational cost makes them unsuitable for real-time systems with strict efficiency demands.

\begin{figure*}[t]
    \centering
    \includegraphics[
    width=0.69\linewidth,
    trim=0 0 1.5mm 0,
    clip
]{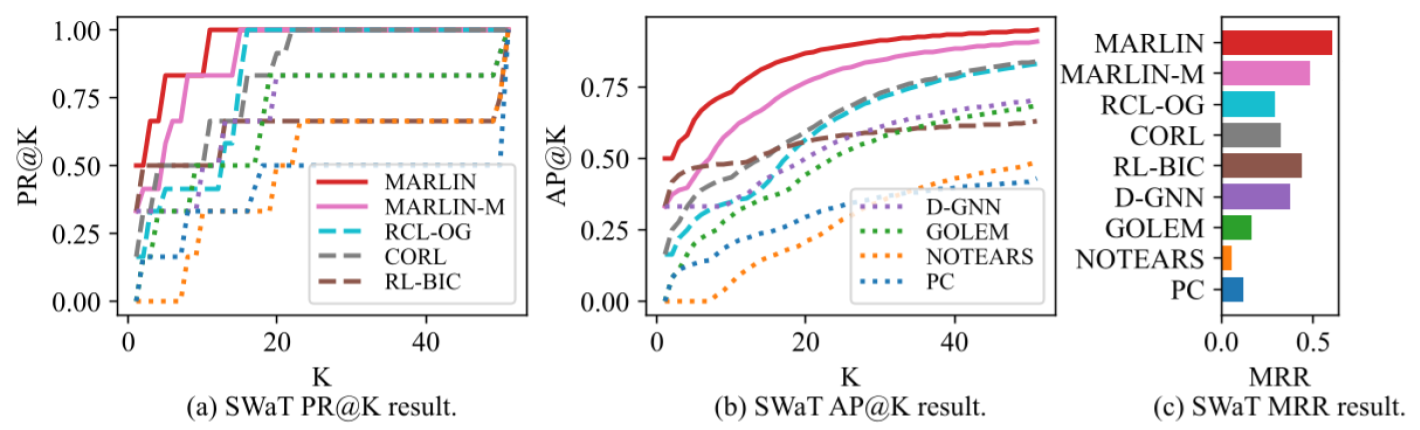}
    \caption{
    Overall performance on the SWaT dataset across (a) PR@$K$, (b) AP@$K$, and (c) MRR metrics.
    }
    \label{fig:result_swat}

\end{figure*}

\subsection{Application to Root Cause Analysis}

To validate \sysname{}'s effectiveness, we evaluate its performance on real data. Due to the lack of ground-truth DAGs in complex real-world data, we assess its incremental DAG learning capability through root cause analysis (RCA), a key task in causal discovery~\cite{wang2023incremental,wang2023hierarchical,wang2023interdependent,zheng2024mulan}. After learning the DAG, we perform random walk with restarts~\cite{tong2006fast} to generate a rank list and compute three widely-used metrics—PR@$K$, AP@$K$, and MRR—where higher values indicate better. Efficiency is measured by the average running time per fault case (ATC). All methods use the same configurations as in the synthetic QR dataset experiment.

\begin{table}[t]
    \scriptsize
    \centering
    \setlength\tabcolsep{2pt}
    \setlength\tabcolsep{1pt}

    \begin{tabular}{l|ccc|ccc|c|c}
        \toprule
         & PR@1 & PR@3 & PR@5 & AP@1 & AP@3 & AP@5 & MRR & ATC$\downarrow$ \\
        \cmidrule(lr){1-1} \cmidrule(lr){2-4} \cmidrule(lr){5-7} \cmidrule(lr){8-8} \cmidrule(lr){9-9}
       PC & 16.7\% & 33.3\% & 50.0\% & 16.7\% & 27.8\% & 36.7\% & 35.2\% & 51 \\
    NOTEARS & 38.9\% & 55.6\% & 72.2\% & 38.9\% & 50.0\% & 57.8\% & 54.4\% & 88 \\
    GOLEM & 11.1\% & 27.8\% & 61.1\% & 11.1\% & 18.5\% & 32.2\% & 30.1\% & 227 \\
    D-GNN & 33.3\% & 50.0\% & 72.2\% & 33.3\% & 40.7\% & 51.1\% & 47.9\% & 315 \\
    RL-BIC & 33.3\% & 44.4\% & 72.2\% & 33.3\% & 40.7\% & 52.2\% & 49.7\% & 171 \\
    CORL & 27.8\% & 72.2\% & 88.9\% & 27.8\% & 51.9\% & 65.6\% & 52.8\% & 141 \\
    RCL-OG & 22.2\% & 77.8\% & \textbf{100\%} & 22.2\% & 51.9\% & 68.9\% & 51.3\% & 122 \\
    \cmidrule(lr){1-1} \cmidrule(lr){2-4} \cmidrule(lr){5-7} \cmidrule(lr){8-8} \cmidrule(lr){9-9}
    \sysname{} & \textbf{61.1\%} & \textbf{94.4\%} & \textbf{100\%} & \textbf{61.1\%} & \textbf{77.8\%} & \textbf{86.7\%} & \textbf{76.4\%} & 63 \\
    \sysnamevariant{} & 44.4\% & 88.9\% & \textbf{100\%} & 44.4\% & 70.4\% & 82.2\% & 67.6\% & \textbf{25} \\

        \bottomrule
    \end{tabular}
    \caption{RCA Performance on OB dataset. All metrics except ATC are better when their values are higher.}
    \label{tab:res_swat}
\end{table}

\subsubsection{Microservice Data.} 
Table \ref{tab:res_swat} shows the RCA performance on the OB dataset. \sysname{} outperforms all baselines in terms of runtime, coming second only to the PC algorithm, demonstrating the success of its intra-batch learning approach and efficiency enhancements. \sysname{} consistently ranks the root cause among the top-3 in nearly all fault cases, indicating its ability to adapt well and learn causal mechanisms amid different system changes. The findings for \sysnamevariant{} are consistent with previous results, significantly enhancing efficiency while learning high-quality DAGs.

\subsubsection{Secure Water Treatment System Data.} 
SWaT and WADI, being larger in scale, pose greater challenges than OB. SWaT results are in Figure~\ref{fig:result_swat}. Non-RL methods struggle with noise and nonlinear causalities, while RL-based methods, though performing better, are too time-consuming for real-time systems. \sysname{} and \sysnamevariant{} surpass all baselines in performance and efficiency, identifying root causes faster with fewer rankings. Their balance between performance and efficiency ensures adaptability. 
Overall, our methods learn high-quality DAGs faster and with less data, even in complex settings.

\begin{table}[t]
    \scriptsize
    \centering
    \setlength\tabcolsep{0.5pt}

    \begin{tabular}{llcccccc}
        \toprule
         & & TPR$\uparrow$  & SHD$\downarrow$ & AUROC$\uparrow$ & SID$\downarrow$ & ATB$\downarrow$\\
        \midrule
        \multirow{2}{*}{$d=20$} & \textbf{\sysname{}} & \textbf{0.99\scriptsize{(0.01)}} & \textbf{5.24\scriptsize{(0.8)}} & \textbf{0.99\scriptsize{(0.00)}} & \textbf{9.6\scriptsize{(2.4)}} & 26\scriptsize{(2)} \\
        & \sysnameablation{} & 0.96\scriptsize{(0.01)} & 14.5\scriptsize{(1.1)} & 0.97\scriptsize{(0.00)} & 22.4\scriptsize{(1.5)} & \textbf{16\scriptsize{(2)}} \\
       \midrule
       
        \multirow{2}{*}{$d=50$} & \textbf{\sysname{}} & \textbf{0.92\scriptsize{(0.02)}} & \textbf{38.9\scriptsize{(8.4)}} & \textbf{0.95\scriptsize{(0.01)}} & \textbf{302\scriptsize{(41.9)}} & \textbf{182\scriptsize{(23)}} \\
        & \sysnameablation{} & 0.88\scriptsize{(0.02)} & 82.1\scriptsize{(11.8)} & 0.90\scriptsize{(0.01)} & 401.6\scriptsize{(46.1)} & 195\scriptsize{(18)} \\
       \midrule
        
        \multirow{2}{*}{$d=100$} & \textbf{\sysname{}} & \textbf{0.85\scriptsize{(0.03)}} & \textbf{105.5\scriptsize{(10.1)}} & \textbf{0.93\scriptsize{(0.01)}} & \textbf{1713.0\scriptsize{(85.9)}} & \textbf{1321\scriptsize{(105)}} \\
        & \sysnameablation{} & 0.82\scriptsize{(0.02)} & 180.6\scriptsize{(6.2)} & 0.86\scriptsize{(0.01)} & 2240.3\scriptsize{(67.7)} & 1687\scriptsize{(111)} \\
    
        \bottomrule
    \end{tabular}
    \caption{Ablation study results on synthetic LG datasets with varying DAG scales ($d={20,50,100}$).}
    \label{tab:ablation_study}
\end{table}

\subsection{Ablation Study}

To further explore the roles of state-specific and state-invariant RL agents in the continuous updating of DAGs, we have designed a single-agent variant, \sysnameablation{}, for an ablation study. \sysnameablation{} utilizes only one RL agent, which takes the current batch data as input and directly learns the complete DAG using the intra-batch learning approach from Section~\ref{sec:intra-batch_learning}, bypassing the disentanglement process.  Table~\ref{tab:ablation_study} presents the empirical results of \sysname{} and \sysnameablation{} on synthetic LG datasets with varying DAG scales.

\sysname{} outperforms \sysnameablation{}, indicating that incremental disentangled DAG learning enhances causal structure learning. For small graphs with simple causal relations, a single agent is more efficient, but as scale and complexity grow, the advantages of a disentangled multi-agent design emerge. \sysname{} rapidly captures state-specific information from new data, whereas \sysnameablation{} adapts more slowly. This underscores the need for both state-specific and state-invariant agents to improve performance and efficiency in fast, incremental DAG learning.

\section{Conclusion}

In this paper, we investigate the challenging problem of learning DAGs in an online setting. We propose \sysname{}, an efficient multi-agent reinforcement learning (RL) framework designed for incremental DAG learning. \sysname{} leverages an efficient intra-batch DAG learning method to learn a policy that maps from a continuous real-valued space to the DAG space. Building on this, \sysname{} incorporates two RL agents—state-specific and state-invariant—to uncover causal relationships and integrates these agents into an incremental learning framework. Additionally, we explore the potential for parallel computation within this framework. Extensive experiments on both synthetic and real-world datasets demonstrate the effectiveness and efficiency of \sysname{} for incremental DAG learning.

\section*{Acknowledgments}
This work done by Yi He has been supported in part by the National Science Foundation (NSF) under Grant Nos. IIS-2505719, IIS-2441449, IOS-2446522, and the Commonwealth Cyber Initiative (CCI).
The work done by Zhong Chen has been supported in part by an Illinois Innovation Network (IIN) sustaining Illinois seed funding grant.
The work done by Chen Zhao has been supported in part by the National Science Foundation (NSF) under Grant No. CNS-2515265.

\bibliography{aaai2026}

\end{document}